\documentclass[10pt,twocolumn,letterpaper]{article}

\usepackage[pagenumbers]{cvpr} 
\usepackage{algorithm}
\usepackage{algorithmic}
\usepackage{multirow}
\usepackage{makecell}
\usepackage{booktabs}
\usepackage{stackengine} 

\usepackage[table]{xcolor} 
\usepackage{mathtools}
\usepackage{graphicx}
\usepackage{nicefrac}
\usepackage[accsupp]{axessibility} 

\definecolor{cvprblue}{rgb}{0.21,0.49,0.74}
\definecolor{lightblue}{rgb}{0.9, 0.95, 1} 
\usepackage[pagebackref,breaklinks,colorlinks,allcolors=cvprblue]{hyperref}

\def\paperID{6618} 
\def\confName{CVPR}
\def\confYear{2026}

\definecolor{myorange}{HTML}{EE8980} 
\definecolor{mypink}{RGB}{255, 150, 200}
\definecolor{myblue}{RGB}{80, 180, 255}
\definecolor{mygray}{HTML}{9CB898}

\def\modelname{UniAVGen}
 
\title{\textcolor{myorange}{Uni}\textcolor{mypink}{A}\textcolor{myblue}{V}\textcolor{mygray}{Gen}: \textcolor{myorange}{Uni}fied \textcolor{mypink}{A}udio and \textcolor{myblue}{V}ideo \textcolor{mygray}{Gen}eration with \\ Asymmetric Cross-Modal Interactions}

\author{%
Guozhen Zhang$^{1,\dagger,*}$ \quad 
Zixiang Zhou$^{2,\dagger}$ \quad
Teng Hu$^{3}$ \quad
Ziqiao Peng$^{4}$ \quad 
Youliang Zhang$^{5}$  \quad \\
Yi Chen$^{2}$ \quad
Yuan Zhou$^{2}$ \quad
Qinglin Lu$^{2}$ \quad
Limin Wang$^{1,6,\ddagger}$\\
$^1$State Key Laboratory for Novel Software Technology, Nanjing University \quad
$^2$Tencent Hunyuan \\
$^3$Shanghai Jiao Tong University \quad 
$^4$Renmin University of China \\
$^5$Tsinghua University \quad
$^6$Shanghai AI Lab \\
\texttt{zgzaacm@gmail.com}\quad\texttt{lmwang@nju.edu.cn}
\\ 
  \newline \textbf{\url{https://mcg-nju.github.io/UniAVGen/}}\\
}

\begin{document}
\twocolumn[{
            \renewcommand\twocolumn[1][]{#1}
            \vspace{-1em}
            \maketitle
            \vspace{-1em}
            \begin{center}
                \vspace{-10pt}
                \centering
                \includegraphics[width=0.9\textwidth]{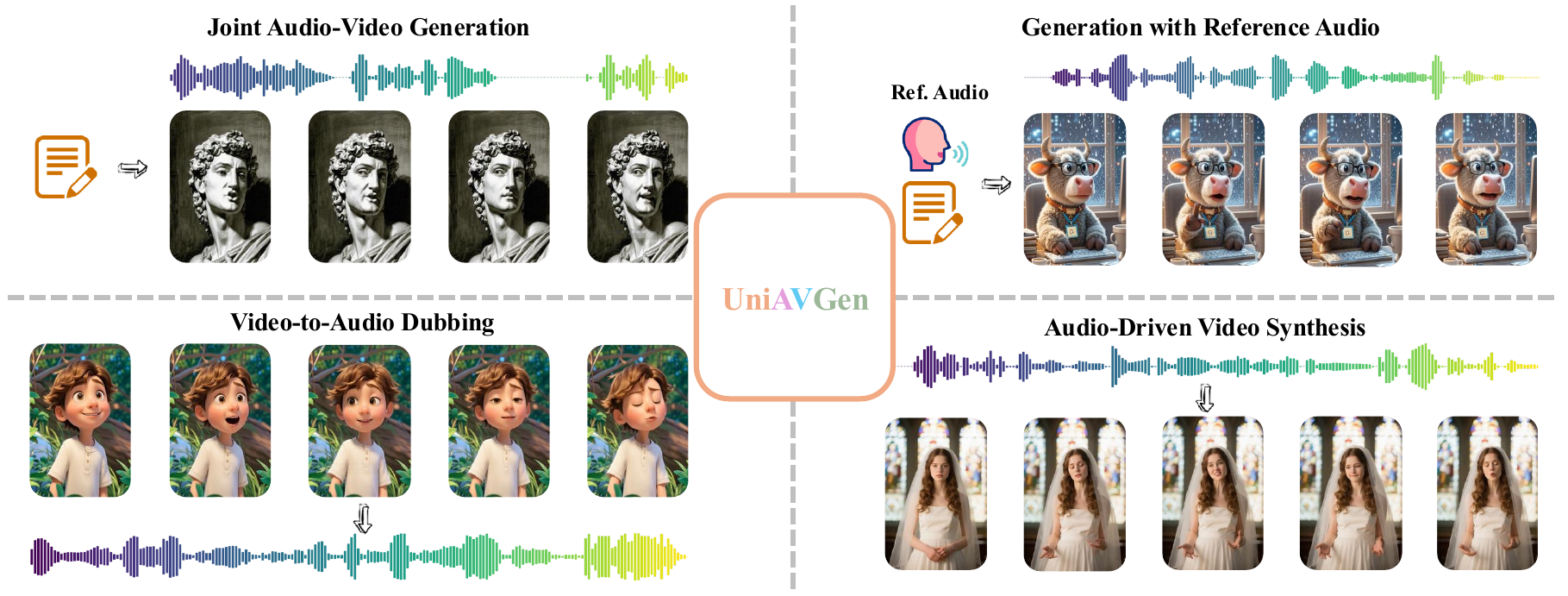}
                \vspace{-1mm}
                \captionof{figure} {
                \textbf{Multi-task compatibility of \modelname.} Leveraging its robust design, \modelname \ can simultaneously tackle  pivotal audio-visual tasks within a single model, eliminating the need for task-specific model designs. 
                }
                \label{fig:teaser}
            \end{center}
        }]

\newcommand\blfootnote[1]{%
\begingroup 
\renewcommand\thefootnote{}\footnote{#1}%
\addtocounter{footnote}{-1}%
\endgroup 
}
\blfootnote{$^*$Work is done during internship at Tencent Hunyuan.}
\blfootnote{$^\dagger$Guozhen Zhang and Zixiang Zhou contribute equally to this work.}
\blfootnote{$^\ddagger$Corresponding author.}
\begin{abstract}

Due to the lack of effective cross-modal modeling, existing open-source audio-video generation methods often exhibit compromised lip synchronization and insufficient semantic consistency.
To mitigate these drawbacks, we propose \textbf{\modelname}, a unified framework for human-centric joint audio and video generation. 
\modelname \ is anchored in a dual-branch joint synthesis architecture, incorporating two parallel Diffusion Transformers (DiTs) to build a cohesive cross-modal latent space. 
At its heart lies an \textbf{Asymmetric Cross-Modal Interaction} mechanism, which enables bidirectional, temporally aligned cross-attention, thus ensuring precise spatiotemporal synchronization and semantic consistency. 
Furthermore, this cross-modal interaction is augmented by a \textbf{Face-Aware Modulation} (FAM) module, which dynamically prioritizes salient regions in the interaction process. 
To enhance generative fidelity during inference, we additionally introduce \textbf{Modality-Aware Classifier-Free Guidance} (MA-CFG), a novel strategy that explicitly amplifies cross-modal correlation signals. 
Notably, \modelname's robust joint synthesis design enables the seamless unification of pivotal audio-visual tasks within a single model. Furthermore, we demonstrate that joint multi-task training can further boost the performance of joint generation.
Comprehensive experiments validate that, with far fewer training samples (1.3M vs. 30.1M), \modelname \ delivers overall advantages in audio-video synchronization, timbre consistency, and emotion consistency.

\end{abstract}    
\section{Introduction}
\label{sec:introduction}

Joint audio-visual generation has emerged as a pivotal trend in state-of-the-art generative AI. Commercial solutions such as Veo3~\cite{veo3_website}, Sora2~\cite{sora2_website}, and Wan2.5~\cite{wan25_website} have achieved exceptional generation fidelity and demonstrated notable practical utility. However, most existing open-source methods~\cite{shan2025hunyuanfoley,cheng2025mmaudio,chen2025hunyuanavatar,hu2025hunyuancustom,lin2025omnihuman,peng2025omnisync} still rely on decoupled pipelines, often leveraging a two-stage paradigm. One paradigm first generates a silent video, then performs separate audio synthesis for post-hoc dubbing~\cite{shan2025hunyuanfoley,cheng2025mmaudio}; the other first generates an audio track to drive subsequent video synthesis~\cite{chen2025hunyuanavatar,hu2025hunyuancustom,lin2025omnihuman}. Regardless of the order, such sequential frameworks inherently suffer from critical limitations: modality decoupling impedes cross-modal interplay during generation, resulting in inadequate semantic consistency and emotional alignment. Consequently, designing effective audio-video alignment in two-stage pipelines grows overly complex, often yielding suboptimal performance.

Recent works have also explored end-to-end joint audio-video generation~\cite{liu2025javisdit,ruan2023mmdiffusion,ishii2024simple,haji2024av,low2025ovi,wang2025universe}. However, existing methods are either confined to generating ambient sounds and fail to synthesize natural human speech~\cite{liu2025javisdit,ruan2023mmdiffusion,ishii2024simple,haji2024av}, or struggle to attain robust audio-visual alignment~\cite{low2025ovi} and produce content lacking fine-grained temporal audio-visual synchronization~\cite{wang2025universe}. Taken together, to date, there remains a lack of highly generalizable and well-aligned audio-video generation method for human-centric joint generation.

To address the aforementioned challenges, we introduce \textbf{\modelname}—a unified framework tailored for joint audio-video generation. We prioritize human-centric audio-video generation not only because this direction remains underexplored in existing works, but also because of its  significant practical utility. 
Specifically, \modelname \ is anchored in a symmetric dual-branch joint synthesis architecture, featuring two parallel Diffusion Transformer (DiT)~\cite{peebles2023scalable,wang2025ddt} streams—one dedicated to video, the other to audio—with identical architectural designs. Crucially, this symmetry establishes representational parity and fosters a cohesive latent space, pivotal for synchronizing joint audio-video generation. To better tackle the intricacies of audio-video alignment for efficient training, we augment this core architecture with three targeted innovations, as detailed below.

First and foremost, at the core of our framework lies an \textbf{Asymmetric Cross-Modal Interaction}  mechanism, which enables bidirectional, temporally aligned cross-modal attention. Equipped with two modal-specifically designed aligners—audio-to-video and video-to-audio—this mechanism injects fine-grained audio semantics into the video stream for precise synchronization, while imparting the temporal dynamics and identity details from video to the audio. To further strengthen this cross-modal synergy and ground it in human-related features, we introduce a \textbf{Face-Aware Modulation} module. Specifically, this component dynamically infers a mask for facial regions using decaying supervision signals, and constrains cross-modal interaction with a gradually relaxed scope. Additionally, to enhance the expressive fidelity of generated content, we propose \textbf{Modality-Aware Classifier-Free Guidance}—a novel strategy that explicitly amplifies cross-modal correlation signals during the classifier-free guidance stage. This targeted enhancement significantly boosts emotional intensity in audio and motion dynamics in video, enhancing the overall realism of the generated content.

As shown in \cref{fig:teaser}, beyond joint generation, our framework can also seamlessly and efficiently adapt to different conditional generation tasks, such as video-to-audio dubbing and audio-driven video synthesis. This versatility enables us to unify pivotal audio-video generation tasks under a single paradigm, eliminating the need for task-specific model designs. \textbf{Furthermore, we experimentally demonstrate that a carefully designed multi-stage training strategy can further boost the performance of joint generation through joint multi-task training.}

Our key contributions are summarized as follows:
\begin{itemize}
\item We present \modelname, a unified audio-video generation framework anchored in a dual-branch joint synthesis architecture and an asymmetric cross-modal interaction mechanism, incorporating modal-specific designs to enhance cross-modal consistency in joint generation.
\item We propose a face-aware modulation module to dynamically constrain the regions of cross-modal interaction for more efficient and aligned cross-modal learning.
\item We present modality-aware classifier-free guidance, a novel strategy that selectively amplifies cross-modal dependencies during inference.
\item Leveraging its robust architectural design, UniAVGen can be seamlessly extended to multiple audio-video generation tasks and demonstrates state-of-the-art performance.
\end{itemize}

\section{Related Work}
\label{sec:related_work}

\begin{figure*}[ht]
    \centering
    \includegraphics[width=0.8\linewidth]{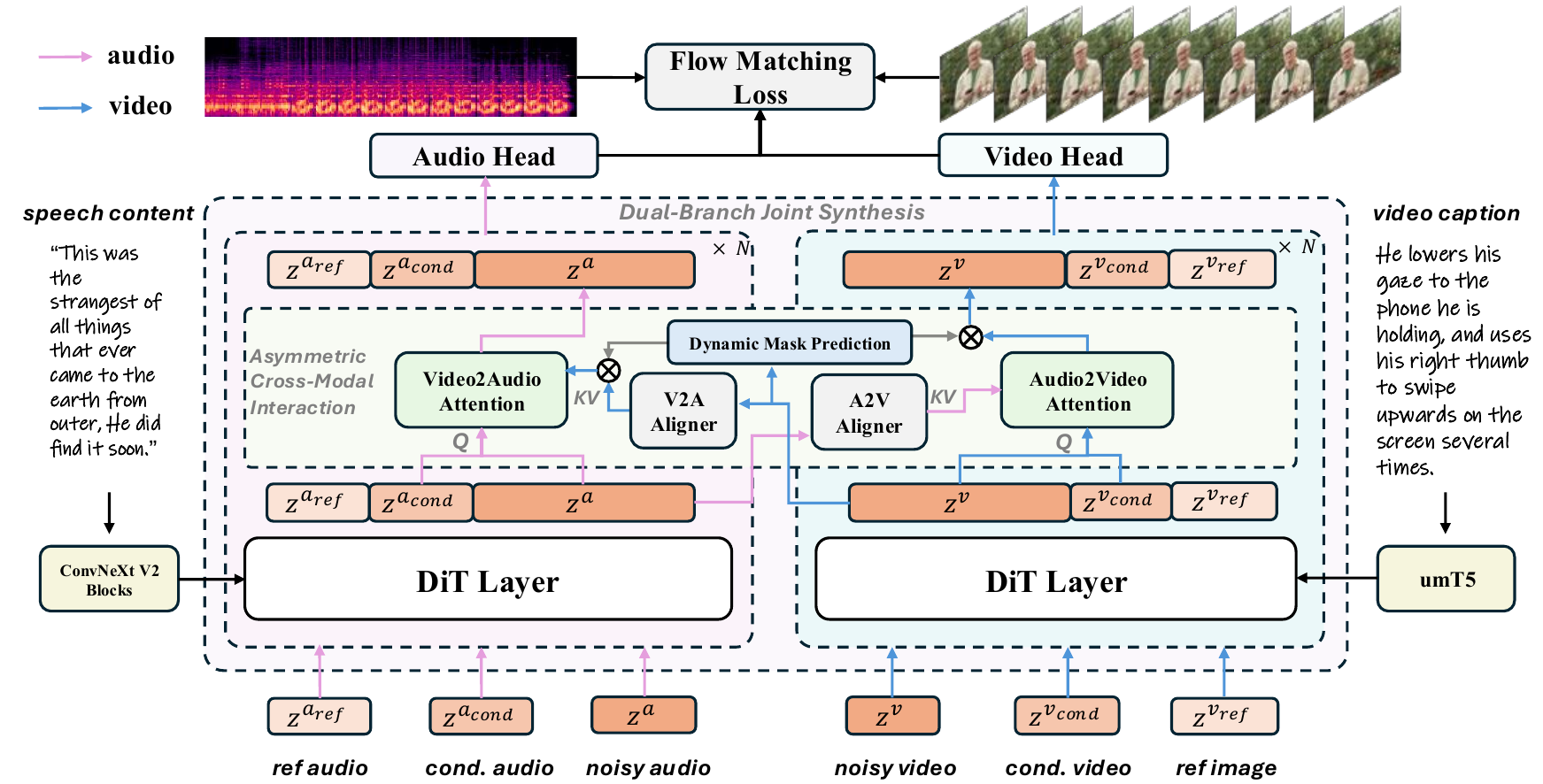}
    \caption{\textbf{Architecture of UniAVGen:} A dual-branch joint synthesis framework with asymmetric cross-modal interaction, augmented by face-aware modulation. Taking a reference image and text prompt as input, it enables coherent audio-video generation.}
    \label{fig:method1}
    \vspace{-5mm}
\end{figure*}

\label{sec:related_work_av_audio}
To enable aligned audio-video generation, the research community has explored three primary paradigms: audio-driven video synthesis, video-to-audio synthesis, and joint audio-video generation.

\noindent\textbf{Audio-driven video synthesis.}
This dominant paradigm typically adopts a two-stage pipeline. First, a Text-to-Speech (TTS) model~\citep{chen2024f5,du2024cosyvoice,peng2024voicecraft,casanova2024xtts,li2025styletts} synthesizes desired audio waveforms from speech content. Subsequently, a separate video synthesis model generates video conditioned on audio, with a focus on lip synchronization~\citep{chen2025hunyuanavatar,lin2025omnihuman,cui2024hallo3,peng2025synctalk++,peng2024synctalk,yariv2024diverse}. While effective for lip synchronization, this cascaded design suffers from inherent modal decoupling: audio is generated without non-verbal cues, leading to poor semantic consistency between audio and video. 

\noindent\textbf{Video-to-audio synthesis.}
The reverse paradigm~\citep{shan2025hunyuanfoley,cheng2025mmaudio,wang2025audiogen,zhang2024foleycrafter,jeong2025read,luo2023diff,polyak2024movie,tian2025audiox,sung2025voicecraft,cong2025flowdubber} aims to generate aligned audio for silent videos. However, it retains two key limitations: first, current methods primarily focus on ambient audio dubbing and lack the ability to synthesize natural human audio; second, it inherits the critical flaw of modal decoupling—videos are generated in an ``auditory vacuum," unaware of the audio they will eventually pair with. 

\noindent\textbf{Joint audio-video generation.}
The most holistic paradigm synthesizes audio and video simultaneously within a single unified framework~\citep{liu2025javisdit,ruan2023mmdiffusion,ishii2024simple,haji2024av,low2025ovi,wang2025universe,zhang2025speakervid,wang2025av,xing2024seeing,yang2024cmmd,tian2025audiox,team2026mova}. Unfortunately, most prior open-source works~\citep{liu2025javisdit,ruan2023mmdiffusion,ishii2024simple,haji2024av, wang2025av,tang2023any} target general joint audio-video generation rather than human specifically—failing to produce high-quality human audio and offering limited practical value. Notably, recent concurrent works—UniVerse-1~\citep{wang2025universe} and Ovi~\citep{low2025ovi}—have begun to support human audio generation. UniVerse-1 stitches two pre-trained audio and video generation models; due to architectural asymmetry, this stitching is complex and yields limited overall performance. Ovi employs a symmetric dual-tower architecture for joint generation, delivering strong performance. However, it lacks modal-specific cross-modal interaction designs and human-specific modulation, resulting in limited generalization in out-of-domain. In contrast, our \modelname{} addresses these gaps by integrating asymmetric cross-modal interaction and face-aware modulation; it thus achieves superior semantic synchronization and robust generalization capabilities.
\section{Method}
\label{sec:method}

Our proposed method, \textbf{\modelname}, is a unified framework for high-fidelity audio-video generation. \modelname \ takes as input a reference speaker image $I^{\text{ref}}$, a video prompt $T^v$ (a caption describing the desired motion or expression), and speech content $T^a$ (the text to be spoken). Additionally, it supports specifying a target voice via an optional reference audio clip $X^{a_{\text{ref}}}$, and enables continuation or conditional generation given reference audio $X^{a_{\text{cond}}}$ and video $X^{v_{\text{cond}}}$. 

\subsection{Overview}

The architecture of \modelname \  is illustrated in \cref{fig:method1}. First, we introduce a dual-branch joint synthesis framework grounded in a symmetric design. For efficient training, we directly adopt the Wan 2.2-5B video generation model~\cite{wang2025wan} as the backbone for the video branch. For the audio branch, we employ the architectural template of Wan 2.1-1.3B—this shares an identical overall structure with Wan 2.2-5B, differing only in the number of channels. This symmetric strategy ensures both branches start with equivalent representational capacity and establishes a natural correspondence between feature maps across all levels. Such structural parity serves as the cornerstone for enabling effective cross-modal interactions, thereby boosting both audio-video synchronization and overall generative quality.

\noindent\textbf{Video branch.}
The video branch operates entirely in the latent space. Specifically, videos are first processed at 16 frames per second and encoded into latent representations $z^v$ using the pre-trained Variational Autoencoder (VAE) from~\cite{wang2025wan}. The reference speaker image $I^{\text{ref}}$ and conditional video are also encoded into latent embeddings $z^{v_{\text{ref}}}$ and $z^{v_{\text{cond}}}$, respectively—with the video branch’s input formed by concatenating these three latent components $z^{\hat v}_t=[z^{v_{\text{ref}}}_0, z^{v_{\text{cond}}}_0, z^v_t]$. For the video caption $T^v$, it is encoded via umT5~\cite{chung2023umT5} into $e^v$, and its embeddings are fed into the Diffusion Transformer (DiT) through cross-attention. Following~\cite{wang2025wan}, we adopt the Flow Matching paradigm~\cite{lipman2022flow}: here, the model $u_{\theta^v}$ is trained to predict the vector field $v_t$, with the training objective formulated as:
\begin{equation}
\mathcal{L}^v = \left\| v_t(z^v_t) - u_{\theta^v}(z^{\hat v}_t, t, e^v) \right\|^2  .
\end{equation}

\begin{figure*}[ht]
    \centering
    \includegraphics[width=0.8\linewidth]{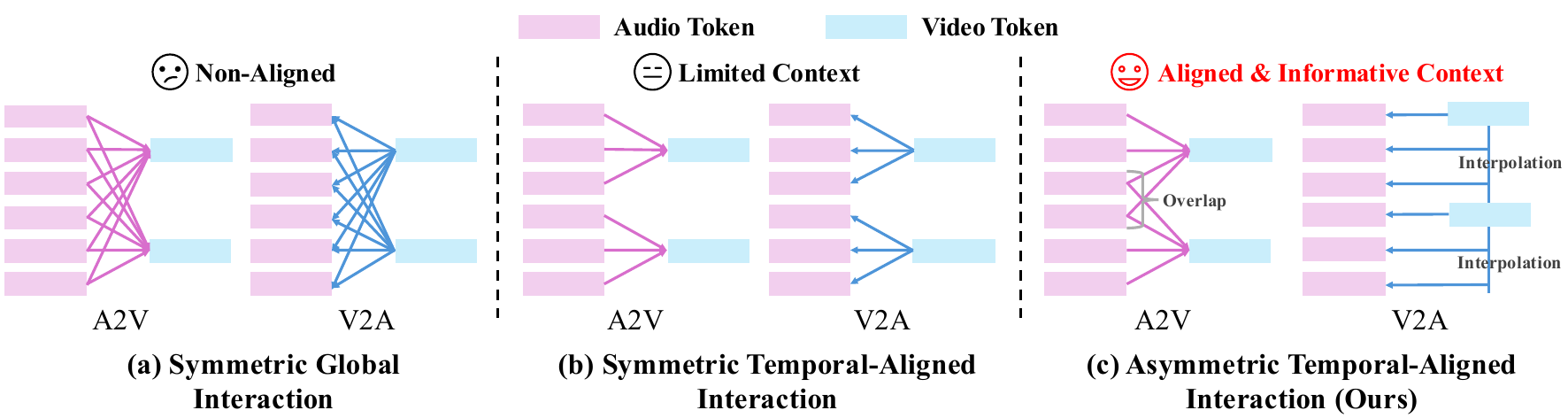}
    \caption{\textbf{Comparison of cross-modal interaction mechanisms}: (a) Global Interaction is simple but poses challenges for convergence; (b) Symmetric Time-Aligned Interaction converges quickly but has limited context utilization; (c) Our Asymmetric Cross-Modal Interaction achieves a superior balance between convergence speed and performance through modal-specific interaction design.}
    \label{fig:method2}
    \vspace{-2mm}
\end{figure*}

\noindent\textbf{Audio branch.}
Following the common practice in text-to-audio (TTS)~\cite{chen2024f5}, audios are first sampled at 24,000 Hz and converted into Mel spectrograms, which serve as the audio latent representation $z^a$. Similarly, the reference audio $X^{a_{\text{ref}}}$ and conditional audio $X^{a_{\text{cond}}}$ are also transformed into their respective latent counterparts $z^{a_{\text{ref}}}$ and $z^{a_{\text{cond}}}$. These three latent components are then concatenated along the temporal dimension to form the audio branch's input $z^{\hat{a}}_t = [z^{a_{\text{ref}}}_0, z^{a_{\text{cond}}}_0, z^a_t]$. The training objective for the audio branch is formulated as:
\begin{equation}
    \mathcal{L}^a = \left\| v_t(z_t^a) - u_{\theta^a}(z^{\hat{a}}_t, t, e^a) \right\|^2 ,
\end{equation}
where $e^a$ denotes the features of the speech content $T^a$ extracted via ConvNeXt~\cite{woo2023convnext} Blocks. These features are further injected into the DiT layers through cross-attention, ensuring the audio generation process is tightly coupled with the acoustic information of the target audio.

\subsection{Asymmetric Cross-Modal Interaction}
\label{subsec:interaction}

While the dual-branch structure establishes structural parity, achieving robust audio-video synchronization demands deep cross-modal interaction. Prior works have primarily employed two designs for this: The first is global interaction~\cite{wang2025universe,low2025ovi}, as shown in \cref{fig:method2}(a), where each token of the current modality interacts with all tokens of the other. While simple, it requires high training costs to converge to strong performance due to lacking explicit temporal alignment. The second is symmetric time-aligned interaction~\cite{ruan2023mmdiffusion}, as shown in \cref{fig:method2}(b), where each video token reciprocally interacts with audio tokens in its corresponding interval. Such methods typically converge faster but access limited contextual information during interaction. To better balance convergence speed and performance, we introduce a novel \textbf{Asymmetric Cross-Modal Interaction} mechanism, comprising two specialized aligners tailored to each modality's unique characteristics.

\noindent\textbf{Audio-to-video (A2V) aligner.}
The A2V aligner ensures precise semantic synchronization by injecting fine-grained audio cues into the video branch. We first reshape the hidden features to align their temporal structure: the video tokens $H^v \in \mathbb{R}^{L^v \times D}$ are reshaped to $\hat{H}^v \in \mathbb{R}^{T \times N^v \times D}$ (where $T$ denotes the number of video latent frames and $N_v$ is the number of spatial tokens per frame), and the audio tokens $H^a \in \mathbb{R}^{L^a \times D}$ are reshaped to $\hat{H}^a \in \mathbb{R}^{T \times N^a \times D}$.

Unlike \cref{fig:method2}(b), we create a \textbf{contextualized} audio representation for each video frame, recognizing that visual articulation is also influenced by preceding and succeeding phonemes. For the $i$-th video latent, we construct an audio context window $C^a_{i}= [ \hat{H}^a_{i-w}, \dots, \hat{H}^a_{i}, \dots, \hat{H}^{a}_{i+w}]$ by concatenating audio tokens from neighboring frames within a window of size $w$. Boundary frames are padded by replicating the features of the first or last frame. Subsequently, we perform frame-wise cross-attention, where the video latent for each frame queries the corresponding contextualized audio latent:
\begin{align}
    \bar{H}^v_i = W_o^v[\hat{H}^v_i + \text{CrossAttention}( Q=W_q\hat{H}^v_i,  \notag\\
     K=W_k^aC^a_{i}, V=W_v^aC^a_{i} )].
\end{align}

\noindent\textbf{Video-to-audio (V2A) aligner.}
Conversely, the V2A aligner aims to embed the audio features with semantics (e.g., timbre, emotion) derived from the visual cues. In A2V, each video latent $i$ maps to a block of $k$ audio tokens.\textbf{ In contrast, for V2A, audio must perceive more precise temporal positional information rather than being confined to a single video latent.} To achieve granular alignment that captures smooth visual transitions, we propose a temporal neighbor interpolation strategy. For each audio token $j$ (corresponding to video latent $i = \lfloor j/k \rfloor$), we compute a unique interpolated video context $C^v_j$—a weighted average of latents from two temporally adjacent video latents: frame $i$ and the subsequent frame $i+1$:
\vspace{-1mm}
\begin{align}
    C^v_j &= (1-\alpha)\hat{H}^v_i + \alpha \hat{H}^v_{i+1}, \notag \\
    \text{where} \quad \alpha &= (j\ \text{mod}\ k) / k.\label{eq:au_context}
\end{align}
For the final block of audio tokens, we simply use $C^v_j = \hat{H}^v_{T-1}$. This interpolated context provides a smooth, time-aware visual signal. Finally, we perform cross-attention where each audio latent queries its corresponding interpolated video context:
\begin{align}
    \bar{H}^a_j = W_o^a\left[\hat{H}^a_j + \text{CrossAttention}\left(Q=W_q^a\hat{H}^a_j, \right. \right. \notag \\
    \left. \left. K=W_k^vC^v_j, V=W_v^vC^v_j\right)\right].\label{eq:v2across}
\end{align}

Finally, $\bar{H}^a$ and $\bar{H}^v$ are reshaped to match the dimensions of $H^a$ and $H^v$, respectively, and injected back as additional features:
\begin{align}
    H^{v} &= H^{v} + \bar{H}^{v},\\
    H^{a} &= H^{a} + \bar{H}^{a}.
\end{align}
To avoid compromising the generative capability of each modality at the start of training, the output matrices $W_o^a$ and $W_o^v$ are both zero-initialized.

\subsection{Face-aware modulation}

For human-centric joint generation, the critical semantic coupling is mostly concentrated in the facial region. Forcing the interaction to process the entire scene is inefficient and risks introducing spurious correlations that destabilize background elements during early training. To address this, we propose a \textbf{Face-Aware Modulation} module that dynamically steers interaction toward the salient regions.

\noindent\textbf{Dynamic mask prediction.}
We introduce a lightweight auxiliary mask-prediction head operating on video features $H^{v_l}$ within each interaction layer $l$ of the denoising network. This head applies layer normalization~\cite{ba2016layer}, a learned affine transformation~\cite{park2019aln}, a linear projection, and sigmoid activation to generate a soft mask $M^l \in (0, 1)^{T \times N_v}$:
\begin{equation}
    M^l = \sigma \left( W_m \left( \gamma \odot \text{LayerNorm}(H^{v_l}) + \beta \right) + b_m \right),
\end{equation}
 where $\odot$ is the element-wise product. To ensure the predicted mask provides a human-aware guide for interaction, we supervise it not only via the final denoising loss but also with an mask loss $
    \lambda^m\mathcal{L}^m = \lambda^m \sum_l \left\| M^l - M^{\text{gt}} \right\|^2$ using the ground-truth face mask $M^{\text{gt}}$~\cite{deng2020retinaface}.
Meanwhile, to avoid over-constraining cross-modal interaction in later training stages, $\lambda^m$ gradually decays to 0 over time. More discussions are provided in \cref{sec:ab3}.

\noindent\textbf{Mask-guided cross-modal interaction.}
The predicted face mask $M^l$ refines cross-modal attention in our asymmetric aligners through two distinct mechanisms: \textbf{(1) A2V interaction:} We employ the mask for selective updates:
\begin{equation}
    H^{v_l} = H^{v_l} + M^{l} \odot \bar{H}^{v_l},
\end{equation}
where $\bar{H}^{v_l}$ denotes the output of A2V cross-attention at layer $l$. This ensures audio information precisely modulates salient regions without disrupting backgrounds during early training. \textbf{(2) V2A interaction:} To enable $M^l$ to strengthen information transfer from the video's salient regions to the audio branch, we modulate the video features $\hat{H}^{v_l}$ as $\hat{H}^{v_l} = M^l \odot \hat{H}^{v_l}$ prior to computing \cref{eq:au_context}.

\subsection{Modality-aware classifier-free guidance}

Classifier-Free Guidance (CFG)~\cite{ho2022classifier} is a cornerstone technique for enhancing conditional fidelity in generative models. However, its conventional design is inherently unimodal. Naively applying it to joint synthesis—where each branch is independently guided by its text prompt—fails to amplify critical cross-modal dependencies. The guidance signal for audio-driven video or video-influenced audio is not explicitly enhanced, limiting the model’s audio-visual synchronization. To address this, we propose \textbf{Modality-Aware Classifier-Free Guidance (MA-CFG)}, a novel scheme that repurposes the guidance mechanism to strengthen cross-modal conditioning. Our key insight is that a single, shared unconditional estimate can serve as the baseline for guiding both modalities simultaneously. This is achieved by performing one forward pass where the conditioning signals for \textit{both} cross-modal interactions are nullified, which is equivalent to unimodal inference.

Specifically, we define the unconditional estimate for the audio and video modalities (without cross-modal interaction) as $u_{\theta_a}$ and $u_{\theta_v}$, and the estimate with cross-modal interaction as $u_{\theta_{a,v}}$. Then, MA-CFG for each modalities can be formulated as:
\begin{align}
    \hat{u}_v &= u_{\theta_v} + s_v  ( u_{\theta_{a,v}}  - u_{\theta_v}), \\
    \hat{u}_a &= u_{\theta_a} + s_a ( u_{\theta_{a,v}} - u_{\theta_a}),
\end{align}
where $s_v$ and $s_a$ are coefficients controlling the guidance strength for the video and audio modalities, respectively.

\begin{figure*}[ht]
    \centering
    \includegraphics[width=0.9\linewidth]{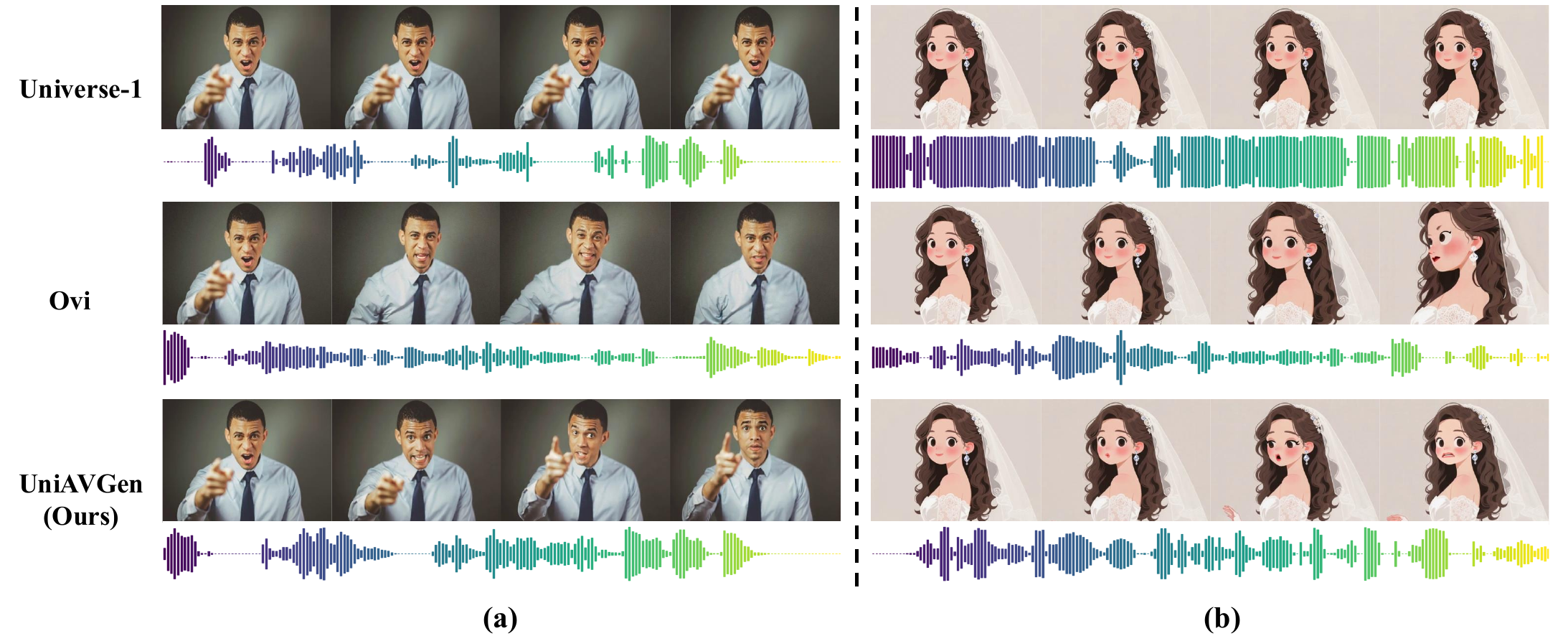}
    \caption{\textbf{Visual comparisons of UniAVGen against concurrent methods Ovi and UniVerse-1.} Specifically, Example (a) uses an in-distribution real human image: UniAVGen and Ovi generate high-fidelity, well-aligned audio-video, while UniVerse-1 is nearly static. Example (b) uses an out-of-distribution (OOD) anime image: Ovi lacks aligned lip/motions (poor generalization), UniVerse-1 stays static with noisy audio; in contrast, our model shows strong generalization, producing coherent, aligned audio-video matching the anime input.}
    \label{fig:exp1}
    \vspace{-2mm}
\end{figure*}

\begin{table*}[t]
    \centering
    \small
    \caption{\textbf{Quantitative comparison with different methods.}}
    \label{tab:method_comparison}
    \resizebox{0.9\textwidth}{!}{
    \begin{tabular}{l c c c c c c c c c c c}
        \toprule
        \multirow{2}{*}{\textbf{Methods}} & \multirowthead{2}[2.5pt]{\textbf{\thead{Joint Training \\ Samples}}} & \multirowthead{2}[2.5pt]{\textbf{\thead{Parameters \\ (S+V)}}} & \multicolumn{3}{c}{\textbf{Audio Quality}} & \multicolumn{3}{c}{\textbf{Video Quality}} & \multicolumn{3}{c}{\textbf{Audio-Video Consistency}} \\
        \cmidrule(lr){4-6} \cmidrule(lr){7-9} \cmidrule(lr){10-12}
        & & & PQ($\uparrow$) & CU($\uparrow$) & WER($\downarrow$) & SC($\uparrow$) & DD($\uparrow$) & IQ($\uparrow$) & LS($\uparrow$) & TC($\uparrow$) & EC($\uparrow$) \\
        \midrule
        \multicolumn{11}{l}{\textit{\footnotesize \textbf{Two-stage Generation}}} \\
        OmniAvatar~\cite{gan2025omniavatar} & - & 21.1B & \textbf{8.15} & \textbf{7.41} & \underline{0.152} & \underline{0.987} & 0.000 & 0.721 &  6.34 &  0.454 & 0.349 \\
        Wan-S2V~\cite{gao2025wan} & - & 16.6B & \textbf{8.15} & \textbf{7.41} & \underline{0.152} & \textbf{0.991} & 0.130 & 0.750 &  \underline{6.35} &  0.481 & 0.375 \\
        \midrule
        \multicolumn{11}{l}{\textit{\footnotesize \textbf{Joint Generation} }} \\
        JavisDiT~\cite{liu2025javisdit} & 10.1M & 3.7B & 5.21 & 3.93 & 0.986 & 0.965 & \underline{0.373} & 0.716 & 1.23 & 0.776 & 0.388 \\
        Universe-1~\cite{wang2025universe} & 6.4M & 7.1B & 4.56 & 4.29 & 0.296 & 0.985 & 0.080 & 0.733 & 1.21 & 0.573 & 0.300 \\
        Ovi~\cite{low2025ovi} & 30.7M & 10.9B & 6.03 & 6.01 & 0.216 & 0.972 & 0.360 & \underline{0.774} &  \textbf{6.48} & \underline{0.828} & \underline{0.558} \\
        \rowcolor{lightblue} UniAVGen & 1.3M & 7.1B & \underline{7.00} & \underline{6.62} & \textbf{0.151} & 0.973 & \textbf{0.410} & \textbf{0.779} & 5.95 & \textbf{0.832} & \textbf{0.573} \\
        \bottomrule
    \end{tabular}
    }
\end{table*}

\subsection{Multi-task unification}
\label{sec:mtu}
As shown in~\cref{fig:method1}, leveraging the symmetry and flexibility of UniAVGen’s overall design, we support multiple input combinations to handle distinct tasks:
(1) \textit{Joint audio-video generation}: The default core task, which takes only text and a reference image as input to generate aligned audio and video.
(2) \textit{Joint generation with reference audio}: Compared to (1), it supports input of a custom reference audio to control the speaker’s timbre. Notably, latents of the reference audio skip cross-modal interaction to preserve the timbre consistency.
(3) \textit{Joint audio-video continuation}: It performs continuation given conditional audio and conditional video. For this task, conditional information also participates in cross-modal interaction to ensure temporal continuity, while its features remain unaffected by interaction to preserve conditional information.
(4) \textit{Video-to-audio dubbing}: When only conditional video is provided to the video branch, the model generates corresponding emotion- and expression-aligned audio based on the video and text. A reference audio can be optionally provided to anchor timbre, and the reference image for the video branch is filled with the first frame of the conditional video.
(5) \textit{audio-driven video synthesis}: When only conditional audio is provided to the audio branch, the model generates expression- and motion-aligned video based on the audio and text.
\textbf{For deeper insights into how multi-task unification facilitates joint generation, we refer the reader to  \cref{sec:ab4}.}
\section{Experiment}
\label{sec:experiment}

\subsection{Implementation details}
UniAVGen is trained in three stages. \textit{Stage 1} focuses on training the audio branch in isolation: here, we only optimize the audio network using its dedicated objective $\mathcal{L}^a$. The training data uses the English subset of the multilingual audio dataset Emilia~\cite{he2024emilia}. We adopt a batch size of 256, a learning rate of $2 \times 10^{-5}$, and the AdamW~\cite{loshchilov2017decoupled} optimizer with parameters $\beta_1 = 0.9$, $\beta_2 = 0.999$, $\epsilon = 1e^{-8}$, for a total of 160k training steps. Once the audio branch achieves robust generative performance, we proceed to \textit{Stage 2}—end-to-end joint training. In this phase, both branches are co-optimized via a composite loss $\mathcal{L}^{\text{joint}} = \mathcal{L}^v + \mathcal{L}^a + \lambda_m \mathcal{L}^m$, where $\lambda_m$ is initialized to 0.1 and decays linearly to 0. The training data here uses an internally collected real human audio-video dataset. We use a batch size of 32, a learning rate of $5e^{-6}$, and the same optimizer settings as Stage 1, for a total of 30k training steps. \textit{Stage 3} involves multi-task learning built on Stage 2, with training configurations consistent with Stage 2. In the training process, the ratio of the 5 tasks mentioned in \cref{sec:mtu} is set to 4:1:1:2:2, with a total of 10k training steps. Inference details are provided in the supplementary materials. 

\subsection{Comparison with previous methods}

\noindent\textbf{Compared methods.}
We select representative methods from two categories of paradigms for comparison: 
(1) \textit{Two-stage Generation}: Since we focus on human=centric joint audio-video generation, we first generate audio using F5-TTS~\cite{chen2024f5}, then generate video from audio with state-of-the-art OmniAvatar~\cite{gan2025omniavatar} and Wan-S2V~\cite{gao2025wan}. 
(2) \textit{Joint Generation}: We select several latest open-source models for comparison: JavisDiT~\cite{liu2025javisdit} focuses on general audio-video joint generation without human audio optimization, UniVerse-1~\cite{wang2025universe} adopts dual pre-trained model stitching, and Ovi~\cite{low2025ovi} employs a symmetric dual-tower architecture with symmetric global cross-model interactions.

\noindent\textbf{Evaluation setting.}
To mitigate test set leakage and better align with the objectives of audio-video generation, we constructed 100 test samples that are not sampled from existing videos. Each sample comprises a reference image, a video caption, and audio content. To comprehensively validate the model's generalization capability—particularly across diverse visual domains—half of these reference images are real-world captures, while the remaining half consists of AIGC-generated content or anime-style visuals.

For evaluation, we measure model performance across three critical dimensions: \textbf{(1) Audio Quality}: Following \cite{wang2025universe}, we adopt AudioBox-Aesthetics~\cite{tjandra2025meta} to evaluate two core metrics: Production Quality (PQ) and Content Usefulness (CU). Additionally, we leverage the Whisper-large-v3~\cite{radford2023robust} model to compute the Word Error Rate (WER) of the generated audio. \textbf{(2) Video Quality}: We utilize VBench~\cite{huang2024vbench}—a widely recognized video evaluation benchmark—to assess video generation quality, focusing on three key metrics: Subject Consistency (SC), Dynamic Degree (DD), and Imaging Quality (IQ). \textbf{(3) Audio-Video Consistency}: Notably, this dimension encompasses three sub-aspects: Lip Synchronization (LS), Timbre Consistency (TC), and Emotion Consistency (EC). Specifically, we employ SyncNet~\cite{chung2016out}'s confidence score to evaluate lip-sync consistency. For timbre and emotion consistency, as no open-source methodologies currently exist to quantify such cross-modal alignment, we instead leverage the multi-modal large language model Gemini-2.5-Pro for evaluation. We set the outputs scores within the range $[0, 1]$. A detailed system prompt (with implementation specifics provided in the supplementary materials) defines the scoring criteria, and the final score for each of these two metrics is computed as the average of three independent evaluations.

\noindent\textbf{Quantitative comparison.}
\cref{tab:method_comparison} summarizes quantitative comparisons between our method and existing baselines: For audio quality, our method demonstrates significant superiority over other joint generation approaches in both acoustic quality and aesthetic metrics, with its WER further outperforming F5-TTS—underscoring stronger alignment with linguistic content. Turning to video quality, while two-stage methods exhibit stronger identity consistency, their dynamism scores are near-zero, reflecting their inability to generate actions congruent with audio-driven emotions; in contrast, our method achieves the highest dynamism and aesthetic quality while retaining identity consistency comparable to state-of-the-art alternatives. Notably, for the critical audio-video consistency metric, our method—despite utilizing the fewest effective training samples—shows clear advantages over competitors in timbre and emotion alignment, while maintaining lip-sync performance on par with leading methods. Such training efficiency is attributed to the proposed asymmetric cross-modal interaction mechanism and face-aware modulation.

\noindent\textbf{Qualitative comparison.}
\cref{fig:exp1} presents visual comparisons of UniAVGen against recent concurrent methods Ovi and UniVerse-1. Specifically, Example (a) uses a real human image aligned with the training distribution: both UniAVGen and Ovi generate high-fidelity audio and videos, with motions and emotions tightly aligned to the audio, whereas UniVerse-1 exhibits near-static behavior. Example (b) employs an anime image—out-of-distribution (OOD) relative to the training set: Ovi fails to produce lip movements and motions aligned with the audio, highlighting its constrained generalization capacity; UniVerse-1 remains static and generates noisy audio. In contrast, our model exhibits robust generalization, generating coherent audio and motions that align with the input anime image.

\begin{table}[t]
    \centering
     \caption{\textbf{Ablation studies on the design of interaction.}}
    \setlength{\tabcolsep}{2.5mm}
    \resizebox{0.35\textwidth}{!}{
    \begin{tabular}{cccccc}
        \toprule
        \multirow{2}{*}{ } & \multicolumn{2}{c}{\textbf{Interaction}} & \multirow{2}{*}{\textbf{LS($\uparrow$)}} & \multirow{2}{*}{\textbf{TC($\uparrow$)}} & \multirow{2}{*}{\textbf{EC($\uparrow$)}} \\
        \cmidrule(lr){2-3} 
        & \textbf{A2V} & \textbf{V2A} &  &  &  \\
        \midrule
        (1) & SGI & SGI & 3.46 & 0.667 & 0.459 \\
        (2) & STI & STI & 3.73 & 0.685 & 0.472 \\
        (3) & STI & ATI & 3.88 & 0.705 & 0.492 \\
        (4) & ATI & STI & 3.97 & 0.691 & 0.483\\
        \rowcolor{lightblue}  (5) & ATI & ATI & \textbf{4.09} & \textbf{0.725} & \textbf{0.504} \\
        \bottomrule
    \end{tabular}
    \vspace{-7mm}
    }
   
    \label{tab:ablation1}
\end{table}
\begin{figure}[t]
    \centering
    \includegraphics[width=0.9\linewidth]{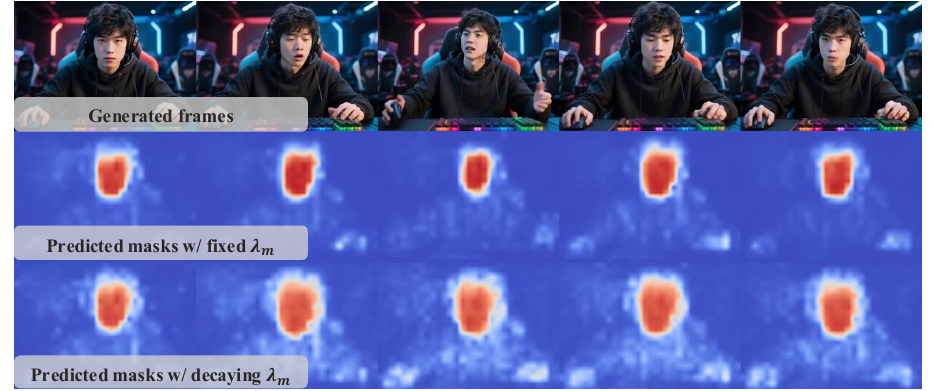}
    \caption{\textbf{Visual comparisons of predicted masks with fixed $\lambda_m$ and decaying $\lambda_m$.} Zoom in for the best view.}
    \label{fig:exp15}
    \vspace{-7mm}
\end{figure}
\subsection{Ablations}
For efficient ablation studies, unless otherwise specified, the following ablation results default to those from the first 10k steps of Stage 2 training. 
The \colorbox{lightblue}{colored background} indicates our default setting.

\subsubsection{Cross-modal interaction design}
As a core architectural component, we perform detailed ablation studies on the design of the cross-modal interaction module, as shown in \cref{tab:ablation1}. Consistent with the three mechanisms depicted in \cref{fig:method2}, this table denotes Symmetric Global Interaction as SGI, Symmetric Temporal-Aligned Interaction as STI, and our proposed Asymmetric Temporal-Aligned Interaction as ATI. SGI exhibits substantial performance deficits compared to STI with the same number of training steps—this confirms that temporal-aligned designs more effectively facilitate model convergence. Relative to STI, our proposed ATI delivers significant improvements in both A2V and V2A tasks: For A2V, ATI more robustly enhances timbre and emotion consistency between audio and video, validating that it indeed strengthens audio's perception of facial expressions and movements across adjacent video frames; for V2A, it further boosts lip synchronization accuracy, confirming that it enables video frames to better capture information from adjacent audio segments. 

\begin{table}[t]
    \centering
    \caption{\textbf{Ablation studies on the face-aware modulation.}}
    \setlength{\tabcolsep}{2.5mm}
    \resizebox{0.8\linewidth}{!}{
    \begin{tabular}{lccc}
        \toprule
        \textbf{Settings} & \textbf{LS($\uparrow$)} & \textbf{TC($\uparrow$)} & \textbf{EC($\uparrow$)} \\
        \midrule
        (a) without FAM & 3.89 & 0.705 & 0.489 \\
        (b) unsupervised FAM & 3.92 & 0.701 & 0.492 \\
        (c) FAM with fixed $\lambda_m$ &  \textbf{4.11} & 0.719 & 0.497 \\
        \rowcolor{lightblue} (d) FAM with decaying $\lambda_m$ & 4.09 & \textbf{0.725} & \textbf{0.504} \\
        \bottomrule
    \end{tabular}
    }
    \vspace{-5mm}
    \label{tab:ablation2}
\end{table}

\subsubsection{Effectiveness of face-aware modulation}
\label{sec:ab3}
\noindent We evaluate the effectiveness of Face-aware Modulation (FAM) through two key analyses. First, to confirm that our lightweight dynamic mask prediction module can reliably localize valid facial regions, we visualize average face masks predicted across layers with fixed $\lambda_m$ in \cref{fig:exp15}. This visualization demonstrates that our module effectively pinpoints face-salient regions. Additionally, when trained with decaying \(\lambda_m\), the predicted masks still effectively capture facial regions while increasing weights on body regions—thereby enhancing the flexibility of cross-modal interactions. To further validate the FAM strategy, we compare performance under four configurations in Table 4: without FAM, unsupervised FAM, FAM with fixed $\lambda_m$, and FAM with decaying $\lambda_m$. Two critical insights emerge: (1) Supervised FAM yields significant improvements in overall audio-video consistency, indicating that constrained masks facilitate training convergence; (2) Decaying loss weights outperform fixed weights, indicating that gradually relaxing constraints on interaction locations during training further enhances the timbre and emotion consistency.

\subsubsection{Modality-aware classifier-free guidance}
\begin{figure}[t]
    \centering
    \includegraphics[width=0.9\linewidth]{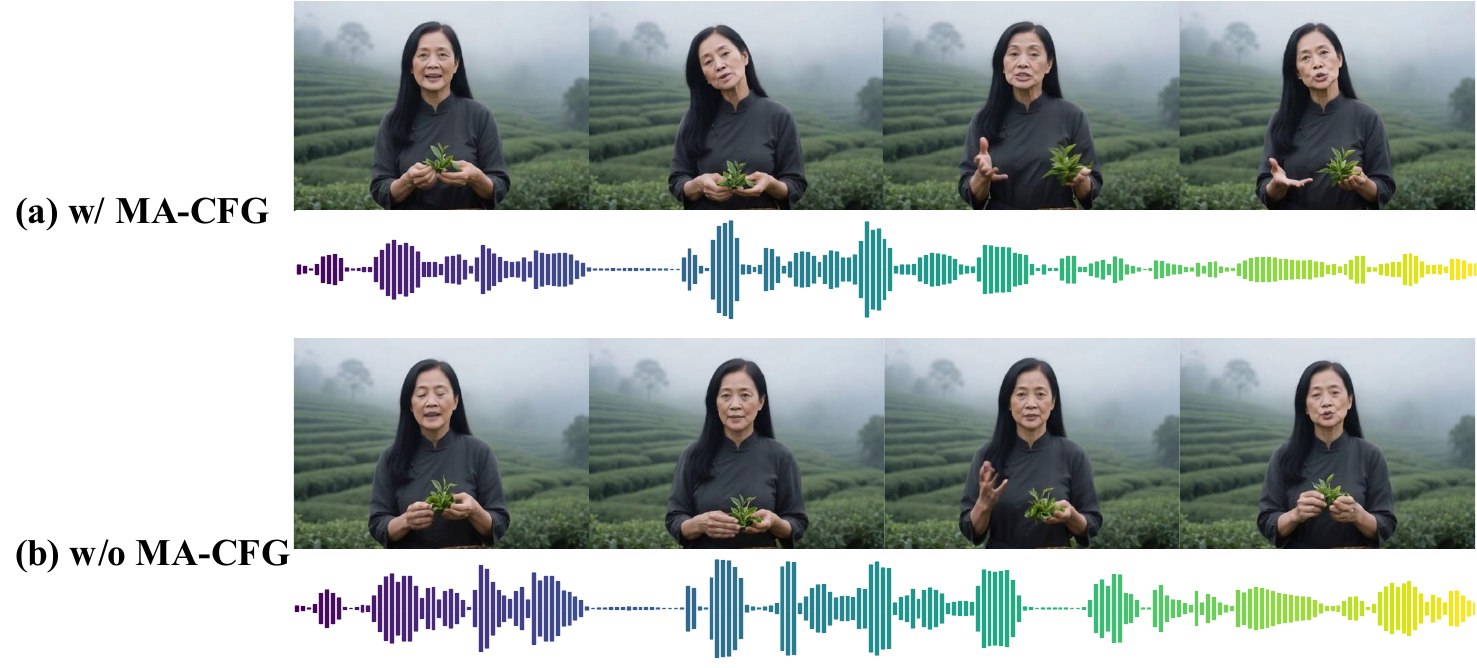}
    \caption{\textbf{Visual comparisons of joint generation results with and without MA-CFG.}     Zoom in for the best view.}
    \label{fig:exp2}
    \vspace{-2mm}
\end{figure}

To demonstrate the effectiveness of MA-CFG, we provide visual comparisons in \cref{fig:exp2}. Without MA-CFG, while audio and video remain generally consistent, the generated character’s emotions and body movements are insufficiently aligned with the audio’s emotional cues. With MA-CFG, by contrast, the jointly generated character exhibits facial expressions and body movements more tightly aligned with audio emotions, alongside more natural lip synchronization.

\begin{figure}[t]
    \centering
    \includegraphics[width=0.75\linewidth]{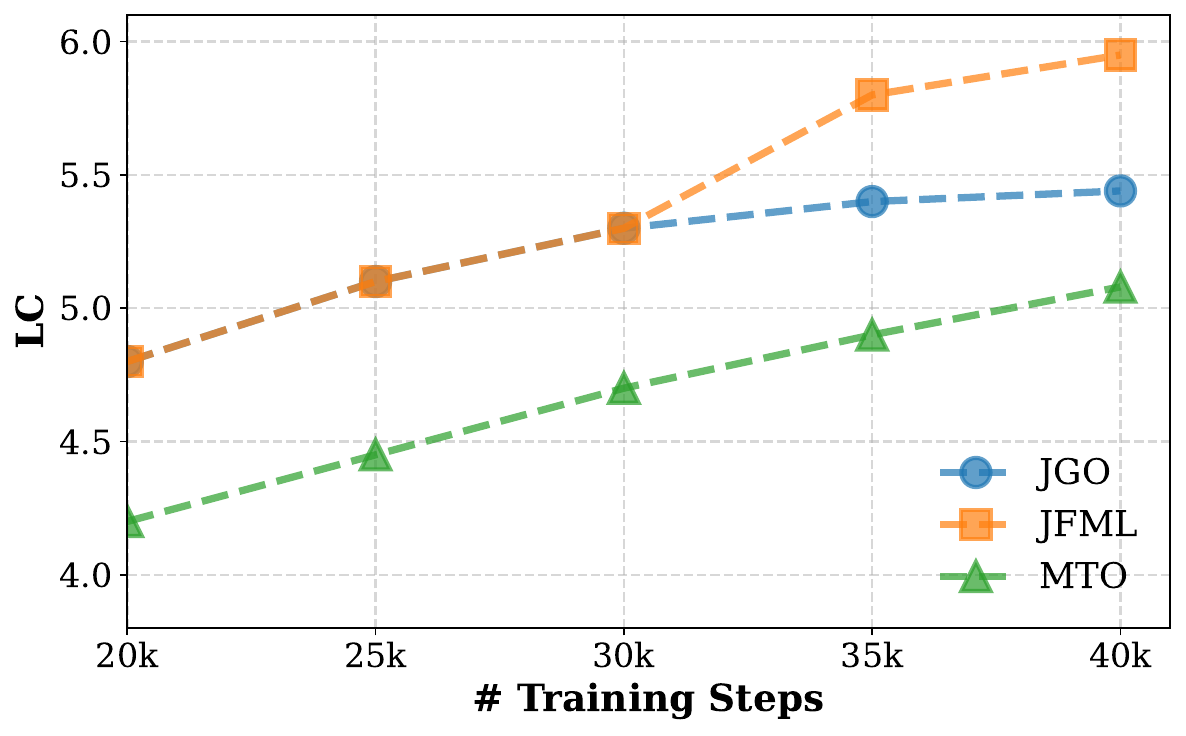}
    \vspace{-3mm}
    \caption{\textbf{Comparisons of different training strategies.}}
    \label{fig:exp3}
    \vspace{-5mm}
\end{figure}

\subsubsection{Analysis of training strategies}
\label{sec:ab4}
As shown in \cref{fig:exp3}, we compare the LC metric of our model under three distinct training strategies: train joint generation only (denoted as JGO), train joint generation first then multi-task learning (denoted as JFML), and multi-task training throughout (denoted as MTO). First, JGO exhibits a lower performance ceiling than JFML, which we attribute to the ability of multi-task joint training to further strengthen cross-modal interaction. For instance, video-to-audio dubbing enhances the audio branch’s capture of conditional information from video, while audio-driven video synthesis deepens the video branch’s perception of the audio branch. Second, MTO demonstrates slower convergence speed than both JGO and JFML. This likely stems from the fact that joint generation is more task-intensive than conditional generation tasks—training the model with conditional tasks from the start may cause it to get trapped in local optima. In contrast, pre-training with joint generation lays a solid foundation for subsequent conditional tasks, allowing JFML to achieve the best overall performance.

\section{Conclusion}

This work introduced UniAVGen, a unified framework for generating high-quality audio and video jointly. At its core lies the asymmetric cross-modal interaction (ATI). Unlike symmetric or global interaction designs, ATI enables modality-specific temporal alignment: it allows audio to efficiently perceive dynamics across adjacent video frames while empowering video frames to capture audio cues from neighboring audio segments. Complementing ATI, we further propose the Face-aware Modulation (FAM) module, which dynamically localizes facial regions and enhances interaction precision. Additionally, we introduce MA-CFG during inference to explicitly strengthen cross-modal influences. Overall, UniAVGen sets a new benchmark for audio-video generation and  paves the way for more practical and versatile multi-modal generation systems.


\paragraph {\bf Acknowledgements.} {\small This work is supported by the Basic Research Program of Jiangsu (No. BK20250009), the Fundamental and Interdisciplinary Disciplines Breakthrough Plan of the Ministry of Education of China (No. JYB2025XDXM118), and the Collaborative Innovation Center of Novel Software Technology and Industrialization.}

{
    \small
    \bibliographystyle{ieeenat_fullname}
    \bibliography{main}
}

\clearpage
\setcounter{page}{1}
\maketitlesupplementary

\section{Additional implementation details}
\subsection{Context window size for A2V aligner}
In the Asymmetric Cross-Modal Interaction mechanism (Sec. 3.2 of the main paper), the audio context window size \( w \) for A2V aligner is set to $\frac{1}{2}$. Specifically, for the \( i \)-th video latent frame, the audio context window \( C_i^a \) concatenates audio tokens from \( i-\frac{1}{2} \) to \( i+\frac{1}{2} \) (i.e., 2 audio segments in total), ensuring sufficient contextual phoneme information for precise lip synchronization. Boundary frames (when \( i-\frac{1}{2} < 0 \) or \( i+\frac{1}{2} \geq T \)) are padded by replicating the first or last audio frame's features to avoid information loss.

\subsection{Temporal alignment in interaction}
Due to the design of existing video VAEs, each video latent, except the first one which corresponds to a single frame, is associated with four consecutive frames. To ensure precise temporal alignment during cross-modal interaction, we explicitly account for this characteristic of video latents. Specifically, for A2V alignment, we first compute the audio window size per frame by dividing the number of audio tokens by the actual number of video frames. We then determine the corresponding audio window for each video latent, meaning the effective window for the first latent is a quarter the size of those for subsequent latents. For V2A alignment, we first upsample the video latents to match audio’s fine-grained temporal resolution: each latent (except the first) is replicated four times. With this temporal alignment, we then compute the video context.

\subsection{Inference details}

We employ the Euler ODE solver with 50 sampling steps and leverage the Vocos vocoder~\cite{siuzdak2023vocos} to convert generated log mel spectrograms into audio signals. For MA-CFG, we empirically set \(s_v=3\) and \(s_a=2\). To stabilize audio-visual quality while using CFG, we further adopt CFG interval~\cite{kynkaanniemi2024applying}, which restricts classifier-free guidance exclusively to the high-frequency generation phase with the interval set to $[0.5, 1]$. Additionally, for efficiency, we set the text condition to empty during unimodal sampling in MA-CFG, which further reinforces text control.

\section{System prompt for evaluation}
We use the following system prompts to evaluate Timbre Consistency (TC) and Emotion Consistency (EC) via Gemini-2.5-Pro. The prompt is designed to ensure objective, reproducible scoring (0-1 scale, 2 decimal places):

{\fontsize{7.6pt}{9pt}\selectfont  
\begin{verbatim}
You are an expert in audio and video understanding. 
Now you will receive an audio and video clip. Plea-
se judge the consistency between the timbre and em-
otion of the audio and video, and give a score bet-
ween 0 and 1.

For timbre evaluation (score a), it is divided into
5 grades based on gender and age matching:
1. 0 points: Completely inconsistent (e.g., video 
shows a woman but audio is a man's voice; age diff-
erence is extremely obvious)
2. 0.25 points: Severely inconsistent (one of gende-
r or age is seriously mismatched, the other has sli-
ght inconsistency)
3. 0.5 points: Partially inconsistent (one of gender
or age is mismatched, the other is consistent)
4. 0.75 points: Basically consistent (gender and age
are roughly matched, with minor details inconsistent)
5. 1 point: Perfectly consistent (gender and age are
completely matched without any differences)

For emotion evaluation (score b), it is divided into
5 grades based on frame-level emotion matching and
body language correspondence:
1. 0 points: No correspondence at all (no frame mat-
ches, body language has nothing to do with audio)
2. 0.25 points: Rarely corresponding (very few frames
match, body language basically does not correspond)
3. 0.5 points: Partially corresponding (about half of
the frames match, body language partially corresponds)
4. 0.75 points: Basically corresponding (most frames 
match, body language roughly corresponds)
5. 1 point: Perfectly corresponding (every frame mat-
ches, body language fits audio perfectly)

You should return the following JSON format:

{"score":[a, b],"reason":"xxx"}

Where a is the timbre score, b is the emotion score,
and reason is the specific reason for the score, wh-
ich should not exceed 100 words.
\end{verbatim}
}
Each sample is evaluated 3 times independently, and the average score is reported.

\begin{table}[t]
\centering
\vspace{-2mm}
\renewcommand{\arraystretch}{1.05}

\caption{\textbf{User study statistics.}}
\vspace{-2mm}
\resizebox{0.42\textwidth}{!}{

    \begin{tabular}{lccc}
    \toprule
    \textbf{Methods} & \textbf{AQ($\uparrow$)} & \textbf{VQ($\uparrow$)} & \textbf{AVC($\uparrow$)}\\
    \midrule
    Universe-1~\cite{wang2025universe} & 2.35\% & 0.00\% & 0.00\% \\
    Ovi~\cite{low2025ovi} & 28.75\% & 37.40\% & 25.70\% \\
    \rowcolor{lightblue} \textbf{UniAVGen (ours)} & \textbf{68.90}\% & \textbf{62.60}\% & \textbf{74.30}\% \\
    \bottomrule
    \end{tabular}

}
\label{tab:user_study}
\end{table}

\section{User study}
A comprehensive user study was also performed to further underscore the advantages of our method. Participants evaluated and selected the top-generated videos by assessing audio quality (AQ), video quality (VQ), and overall audio-visual coherence (AVC). Results from 34 participants, presented in \cref{tab:user_study}, reveal that our approach achieves superior overall audio-visual quality and enhanced consistency between audio and video compared to recent methods.

\begin{table}[t]
\centering
\vspace{-2mm}
\renewcommand{\arraystretch}{1.05}
\caption{\textbf{Results on GRID~\cite{cooke2006audio} under Dubbing Setting 3.0.}}
\vspace{-2mm}
\resizebox{0.42\textwidth}{!}{

    \begin{tabular}{lccc}
    \toprule
    \textbf{Methods} & \textbf{LSE-C($\uparrow$)} & \textbf{LSE-D($\downarrow$)} & \textbf{WER($\downarrow$)}\\
    \midrule
    StyleDubber~\cite{cong2024styledubber} & 5.94 & 9.75 & 15.40 \\
    EmoDubber~\cite{cong2025emodubber} & 7.25 & 6.83 & 14.72 \\
    \rowcolor{lightblue} \textbf{UniAVGen (ours)} & \textbf{7.59} & \textbf{6.11} & \textbf{10.64} \\
    \bottomrule
    \end{tabular}

}
\vspace{-2mm}
\label{tab:v2a}
\end{table}

\begin{table}[t]
\centering

\caption{\textbf{Results on EMTD~\cite{meng2025echomimicv2}.}}
\vspace{-2mm}
\renewcommand{\arraystretch}{1.05}
\resizebox{0.44\textwidth}{!}{
    \begin{tabular}{lcccc}
    \toprule
    \textbf{Methods} &  \textbf{LSE-C($\uparrow$)} & \textbf{LSE-D($\downarrow$)} & \textbf{FID($\downarrow$)} & \textbf{FVD($\downarrow$)}\\
    \midrule
    OmniAvatar~\cite{gan2025omniavatar} & 7.19 & 6.90 & 45.02 & 459.44 \\
    Wan-S2V~\cite{gao2025wan} & \textbf{7.24} & 6.92 & 44.02 & \textbf{451.44} \\
    \rowcolor{lightblue} \textbf{UniAVGen (ours)} & 7.05 & \textbf{6.85} & \textbf{43.97} & 469.85 \\
    \bottomrule
    \end{tabular}

}
\label{tab:a2v}
\end{table}

\section{Evaluation on conditional tasks}

While UniAVGen is primarily designed for high-quality joint audio-visual generation, we further evaluate its performance on public benchmarks of other conditional generation tasks after multi-task joint training to ensure the completeness of this work. For video-to-audio dubbing, we test on the widely used GRID benchmark~\cite{cooke2006audio} with three metrics: LSE-C~\cite{chung2016out}, LSE-D~\cite{chung2016out} and WER. As shown in \cref{tab:v2a}, we compare performance under Dubbing Setting 3.0, which adopts unseen speakers as reference audio. Without complex or task-specific designs, our model achieves superior consistency and lower WER. For audio-to-video synthesis, we utilize the half-body animation benchmark EMTD~\cite{meng2025echomimicv2} and compare against state-of-the-art audio-driven models~\cite{gan2025omniavatar,gao2025wan}. As presented in \cref{tab:a2v}, our model attains near-SOTA performance with only simple multi-task fine-tuning. These results further validate the practicality and generalization capability of UniAVGen.

\section{Extended ablation studies}
We supplement additional ablation experiments to validate the robustness of our core designs:

\subsection{Exploration of interaction  insertion positions}
Rationally integrating the interaction module is another critical consideration, which we address from two perspectives. First, at the layer-level (see \cref{tab:abi1}), we explore four schemes: inserting into all layers, the first half of layers, the last half of layers, and interleaved insertion. Interleaved insertion yields the best results, indicating that appropriate yet not excessive cross-modal interaction better enhances the stability of multi-modal learning. Second, at the operation-level: built on the DiT architecture of Wan2.2-5b, each DiT block comprises self-attention, text cross-attention, and FFN. We ablate the module insertion at three distinct positions: 1) before self-attention, 2) before cross-attention, and 3) before FFN. As shown in \cref{tab:abi2}, position 1) achieves the optimal performance, suggesting that fully preserving the operational flow of each block facilitates better inheritance of pretrained capabilities. 

\begin{table}[t]
    \centering
    \caption{\textbf{Ablation studies on the layer-level insertion.}}
    \vspace{-2mm}
    \setlength{\tabcolsep}{2.5mm}
    \resizebox{0.75\linewidth}{!}{
    \begin{tabular}{lccc}
        \toprule
        \textbf{Settings} & \textbf{LS($\uparrow$)} & \textbf{TC($\uparrow$)} & \textbf{EC($\uparrow$)} \\
        \midrule
        (a) all layers & 4.01 & 0.713 & 0.497 \\
        (b) first half of layers & 4.02 & 0.719 & 0.500 \\
        (c) last half of layers &  3.79 & 0.710 & 0.493 \\
        \rowcolor{lightblue} \textbf{(d) interleaved layers} & \textbf{4.09} & \textbf{0.725} & \textbf{0.504} \\
        \bottomrule
    \end{tabular}
    }
    \vspace{-2mm}
    \label{tab:abi1}
\end{table}

\begin{table}[t]
    \centering
    \caption{\textbf{Ablation studies on the operation-level insertion.}}
    \vspace{-2mm}
    \setlength{\tabcolsep}{2.5mm}
    \resizebox{0.75\linewidth}{!}{
    \begin{tabular}{lccc}
        \toprule
        \textbf{Settings} & \textbf{LS($\uparrow$)} & \textbf{TC($\uparrow$)} & \textbf{EC($\uparrow$)} \\
        \midrule
        1) before FFN & 3.85 & 0.715 & 0.490 \\
        2) before cross-attention & 3.98 & 0.721 & 0.499 \\
        \rowcolor{lightblue}  \textbf{3) before self-attention} & \textbf{4.09} & \textbf{0.725} & \textbf{0.504} \\
        \bottomrule
    \end{tabular}
    }
    \vspace{-2mm}
    \label{tab:abi2}
\end{table}

\subsection{Validation of MA-CFG's effectiveness}
As shown in \cref{tab:abcfg}, we compare the performance of four testing strategies: no CFG, vanilla CFG, MA-CFG, and MA-CFG with the interval $[0.5, 1]$. While vanilla CFG improves image quality, its enhancement on modal consistency is negligible. In contrast, MA-CFG significantly boosts audio-visual alignment metrics but slightly degrades image quality. By incorporating the constrained CFG interval, MA-CFG achieves simultaneous improvements in both image quality and modal alignment.

\begin{table}[t]
    \centering
    \caption{\textbf{Ablation studies on the MA-CFG.}}
    \vspace{-2mm}
    \setlength{\tabcolsep}{2.5mm}
    \resizebox{0.9\linewidth}{!}{
    \begin{tabular}{lcccc}
        \toprule
        \textbf{Settings} & \textbf{LS($\uparrow$)} & \textbf{TC($\uparrow$)} & \textbf{EC($\uparrow$)} & \textbf{IQ($\uparrow$)} \\
        \midrule
        (a) no CFG & 5.75 & 0.821 & 0.553 & 0.760 \\
        (b) vanilla CFG & 5.81 & 0.824 & 0.562 & 0.778 \\
        (c) MA-CFG &  \textbf{6.29} & \textbf{0.841} & \textbf{0.580} & 0.752 \\
        \rowcolor{lightblue} \textbf{(d) MA-CFG under $t\in[0.5, 1]$} & 5.95 & 0.832 & 0.573 & \textbf{0.779} \\
        \bottomrule
    \end{tabular}
    }
    \vspace{-2mm}
    \label{tab:abcfg}
\end{table}

\section{Limitations}
Currently, while UniAVGen performs well in speech-video generation, it lacks video-aligned ambient sound generation. Additionally, its ability to generate audio for multi-person scenarios remains constrained by the inflexible text encoder. For future efforts, we will first collecting more general high-quality audio-video data. Meanwhile, we plan to enhance the text encoder of the audio branch, specifically by adopting multi-modal large language models like Qwen-Omni3~\cite{xu2025qwen3}, to enable multi-person scenarios generation. 
\end{document}